\documentclass[preprint,12pt]{elsarticle}

\usepackage{tikz}
\usetikzlibrary{backgrounds}
\usepackage{graphicx}
\usepackage{bm}
\usepackage{amsmath}
\usepackage{setspace}
\usepackage{hyperref}
\usepackage{multirow}
\hypersetup{
    colorlinks=true,       % false: boxed links; true: colored links
}
\usepackage{amssymb}
\usepackage{lineno}

\journal{Journal Name}

\begin{document}

\begin{frontmatter}


\title{Hidden Physics Models: Machine Learning of Nonlinear Partial Differential Equations}



%\author{Maziar Raissi$^{1}$, Paris Perdikaris$^{2}$, and George Em Karniadakis$^{1}$}
\author{Maziar Raissi and George Em Karniadakis}
%\author{Maziar Raissi}
%\address{$^{1}$Division of Applied Mathematics, Brown University,\\ Providence, RI, 02912, USA\\
%$^{2}$Department of Mechanical Engineering,\\ Massachusetts Institute of Technology,\\ Cambridge, MA, 02139, USA}
\address{Division of Applied Mathematics, Brown University,\\ Providence, RI, 02912, USA}

\begin{abstract}
While there is currently a lot of enthusiasm about ``big data", useful data is usually ``small" and expensive to acquire. In this paper, we present a new paradigm of learning partial differential equations from {\em small} data. In particular, we introduce \emph{hidden physics models}, which are essentially data-efficient learning machines capable of leveraging the underlying laws of physics, expressed by time dependent and nonlinear partial differential equations, to extract patterns from high-dimensional data generated from experiments. The proposed methodology may be applied to the problem of learning, system identification, or data-driven discovery of partial differential equations. Our framework relies on Gaussian processes, a powerful tool for probabilistic inference over functions, that enables us to strike a balance between model complexity and data fitting. The effectiveness of the proposed approach is demonstrated through a variety of canonical problems, spanning a number of scientific domains, including the Navier-Stokes, Schr\"{o}dinger, Kuramoto-Sivashinsky, and time dependent linear fractional equations. The methodology provides a promising new direction for harnessing the long-standing developments of classical methods in applied mathematics and mathematical physics to design learning machines with the ability to operate in complex domains without requiring large quantities of data.

% {\color{red}{We demonstrate the method by considering ... examples ... say something about accuracy, UQ, scalability to big domains, long-term integration.}}
\end{abstract}


\end{frontmatter}

% \linenumbers

\section{Introduction}
There are more than a trillion sensors in the world today and according to some estimates there will be about 50 trillion cameras worldwide within the next five years, all collecting data either sporadically or around the clock. However, in scientific experiments, quality and error-free data is not easy to obtain -- e.g., for system dynamics characterized by bifurcations and instabilities, hysteresis, and often irreversible responses. Admittedly, as in all everyday applications, in scientific experiments too, the volume of data has increased substantially compared to even a decade ago but analyzing big data is expensive and time-consuming. Data-driven methods, which have been enabled in the past decade by the availability of sensors, data storage, and computational resources, are taking center stage across many disciplines of science. We now have highly scalable solutions for problems in object detection and recognition, machine translation, text-to-speech conversion, recommender systems, and information retrieval. All of these solutions attain state-of-the-art performance when trained with large amounts of data. However, purely data driven approaches for machine learning present difficulties when the data is scarce relative to the complexity of the system. Hence, the ability to learn in a sample-efficient manner is a necessity in these data-limited domains. Less well understood is how to leverage the underlying physical laws and/or governing equations to extract patterns from small data generated from highly complex systems. In this work, we propose a modeling framework that enables blending conservation laws, physical principles, and/or phenomenological behaviors expressed by partial differential equations with the datasets available in many fields of engineering, science, and technology. This paper should be considered a direct continuation of a preceding one \cite{raissi2017numerical} in which we addressed the problem of inferring solutions of time dependent and nonlinear partial differential equations using noisy observations. Here, a similar methodology is employed to deal with the problem of learning, system identification, or data-driven discovery of partial differential equations \cite{Rudye1602614}.

\section{Problem Setup}
Let us consider parametrized and nonlinear partial differential equations of the general form
\begin{eqnarray}\label{eq:PDE}
&&h_t + \mathcal{N}_x^\lambda h = 0,\ x \in \Omega, \ t\in[0,T],
\end{eqnarray}
where $h(t,x)$ denotes the latent (hidden) solution, $\mathcal{N}_x^\lambda$ is a nonlinear operator parametrized by $\lambda$, and $\Omega$ is a subset of $\mathbb{R}^D$. As an example, the one dimensional Burgers' equation corresponds to the case where $\mathcal{N}_x^\lambda h = \lambda_1 h h_x - \lambda_2 h_{xx}$ and $\lambda = (\lambda_1, \lambda_2)$. Here, the subscripts denote partial differentiation in either time or space. Given noisy measurements of the system, one is typically interested in the solution of two distinct problems. The first problem is that of inference or filtering and smoothing, which states: given fixed model parameters $\lambda$ what can be said about the unknown hidden state $h(t,x)$ of the system? This question is the topic of a preceding paper \cite{raissi2017numerical} of the authors in which we introduce the concept of \emph{numerical Gaussian processes} and address the problem of inferring solutions of time dependent and nonlinear partial differential equations using noisy observations. The second problem is that of learning, system identification, or data driven discovery of partial differential equations \cite{Rudye1602614} stating: what are the parameters $\lambda$ that best describe the observed data? Here we assume that all we observe are two snapshots $\{\bm{x}^{n-1}, \bm{h}^{n-1}\}$ and $\{\bm{x}^{n}, \bm{h}^{n}\}$ of the system at times $t^{n-1}$ and $t^n$, respectively, which are $\Delta t = t^n - t^{n-1}$ apart. The main assumption is that $\Delta t$ is small enough so that we can apply the backward Euler time stepping scheme\footnote{For a general treatment of arbitrary linear multi-step methods as well as Runge-Kutta time stepping schemes we would like to refer the readers to \cite{raissi2017numerical}.} to equation (\ref{eq:PDE}) and obtain the discretized equation 
\begin{equation}\label{eq:BackwardEuler}
h^{n} + \Delta t \mathcal{N}^\lambda_x h^n = h^{n-1}.
\end{equation}
Here, $h^n(x) = h(t^n,x)$ is the hidden state of the system at time $t^n$. Approximating the nonlinear operator on the left-hand-side of equation (\ref{eq:BackwardEuler}) by a linear one we obtain
\begin{equation}\label{eq:BackwardEulerLinearized}
\mathcal{L}^\lambda_x h^n = h^{n-1}.
\end{equation}
For instance, the nonlinear operator
\[
h^n + \Delta t \mathcal{N}_x^\lambda h^n = h^n + \Delta t (\lambda_1 h^{n} h^n_x - \lambda_2 h^n_{xx}),
\]
involved in the Burgers' equation can be approximated by the linear operator
\[
\mathcal{L}_x^\lambda h^n = h^n + \Delta t (\lambda_1 h^{n-1} h^n_x - \lambda_2 h^n_{xx}),
\]
where $h^{n-1}(x)$ is the state of the system at the previous time $t^{n-1}$.

\section{The Basic Model}
Similar to Raissi et al. \cite{Raissi2017736, raissi2017machine}, we build upon the analytical property of Gaussian processes that the output of a linear system whose input is Gaussian distributed is again Gaussian. Specifically, we proceed by placing a Gaussian process\footnote{Gaussian processes (see \cite{Rasmussen06gaussianprocesses, murphy2012machine}) provide a flexible prior distribution over functions and enjoy analytical tractability. They can be viewed as a prior on one-layer feed-forward Bayesian neural networks with an infinite number of hidden units \cite{neal2012bayesian}. Gaussian processes are among a class of methods known as kernel machines (see \cite{vapnik2013nature, scholkopf2002learning, tipping2001sparse}) and are analogous to regularization approaches (see \cite{tikhonov1963solution, Tikhonov/Arsenin/77, poggio1990networks}).} prior over the latent function $h^{n}(x)$; i.e.,
\begin{equation}\label{eq:Prior}
h^n(x) \sim \mathcal{GP}(0, k(x,x',\theta)).
\end{equation}
Here, $\theta$ denotes the hyper-parameters of the covariance function $k$. Without loss of generality, all Gaussian process priors used in this work are assumed to have a squared exponential\footnote{From a theoretical point of view, each kernel (i.e., covariance function) gives rise to a Reproducing Kernel Hilbert Space (RKHS) \cite{aronszajn1950theory, saitoh1988theory, berlinet2011reproducing} that defines a class of functions that can be represented by this kernel. In particular, the squared exponential covariance function implies smooth approximations. For a more systematic treatment of the kernel-selection problem we would like to refer the readers to \cite{duvenaud2013structure, grosse2012exploiting, malkomes2016bayesian}. Furthermore, more complex function classes can be accommodated by employing nonlinear warping of the input space to capture discontinuities \cite{calandra2016manifold, raissi2016deep}.} covariance function, i.e.,
\[
k(x,x';\theta) = \gamma^2 \exp\left(-\frac12\sum_{d=1}^D w_{d}^2(x_d - x'_d)^2\right),
\]
where $\theta = \left(\gamma,w_1, \cdots, w_D\right)$ are the hyper-parameters and $x$ is a $D$-dimensional vector. The Gaussian process prior assumption (\ref{eq:Prior}) along with equation (\ref{eq:BackwardEulerLinearized}) enable us to capture the entire structure of the operator $\mathcal{L}_x^{\lambda}$ in the resulting multi-output Gaussian process
\begin{equation}\label{eq:HPM}
\begin{bmatrix}
h^{n} \\ 
h^{n-1}
\end{bmatrix}
\sim \mathcal{GP}\left(0, \begin{bmatrix}
k^{n,n} & k^{n,n-1}\\ 
k^{n-1,n} & k^{n-1,n-1}
\end{bmatrix}
\right).
\end{equation}
It is worth highlighting that the parameters $\lambda$ of the operators $\mathcal{L}^\lambda_x$ and $\mathcal{N}^\lambda_x$ turn into hyper-parameters of the resulting covariance functions. The specific forms of the kernels\footnote{It should be noted that for all examples studied in this work the kernels are generated at the push of a button using Wolfram Mathematica, a mathematical symbolic computation program.}
\begin{align}
&k^{n,n}(x,x';\theta), &&k^{n,n-1}(x,x';\theta,\lambda),\nonumber\\
&k^{n-1,n}(x,x';\theta,\lambda), &&k^{n-1,n-1}(x,x';\theta, \lambda),\nonumber
\end{align}
are direct functions of equation (\ref{eq:BackwardEulerLinearized}) as well as the prior assumption (\ref{eq:Prior}); i.e.,
\begin{align}
&k^{n,n} = k, &&k^{n,n-1} = \mathcal{L}_{x'}^\lambda k,\nonumber\\
&k^{n-1,n} = \mathcal{L}_x^\lambda k, &&k^{n-1,n-1} = \mathcal{L}_{x}^\lambda \mathcal{L}_{x'}^\lambda k,\nonumber
\end{align}
We call the multi-output Gaussian process (\ref{eq:HPM}) a \emph{hidden physics model}, because its matrix of covariance functions explicitly encodes the underlying laws of physics expressed by equations (\ref{eq:PDE}) and (\ref{eq:BackwardEulerLinearized}).

\section{Learning}\label{sec:Learning}
Given the noisy data $\{\bm{x}^{n-1}, \bm{h}^{n-1}\}$ and $\{\bm{x}^{n}, \bm{h}^{n}\}$ on the latent solution at times $t^{n-1}$ and $t^n$, respectively, the hyper-parameters $\theta$ of the covariance functions and more importantly the parameters $\lambda$ of the operators $\mathcal{L}_x^\lambda$ and $\mathcal{N}_x^\lambda$ can be learned by employing a Quasi-Newton optimizer L-BFGS \cite{liu1989limited} to minimize the negative log marginal likelihood \cite{Rasmussen06gaussianprocesses}
\begin{equation}\label{eq:NLML}
-\log p(\bm{h}| \theta, \lambda, \sigma^2) = \frac{1}{2} \bm{h}^{T}\bm{K}^{-1}\bm{h} + \frac{1}{2}\log |\bm{K}| + \frac{N}{2} \log (2\pi),
\end{equation}
where $\bm{h} = \begin{bmatrix}
\bm{h}^n \\ 
\bm{h}^{n-1}
\end{bmatrix}
$, $p(\bm{h} | \theta, \lambda, \sigma^2) = \mathcal{N}\left(\bm{0}, \bm{K}\right)$, and $\bm{K}$ is given by
\[
\bm{K} = \begin{bmatrix}
k^{n,n}(\bm{x}^{n},\bm{x}^{n}) & k^{n,n-1}(\bm{x}^{n},\bm{x}^{n-1})\\ 
k^{n-1,n}(\bm{x}^{n-1},\bm{x}^{n}) & k^{n-1,n-1}(\bm{x}^{n-1},\bm{x}^{n-1})
\end{bmatrix} + \sigma^2 \bm{I}.
\]
Here, $N$ is the total number of data points in $\bm{h}$. Moreover, $\sigma^2$ is included to capture the noise in the data and is also learned by minimizing the negative log marginal likelihood. The implicit underlying assumption is that $\bm{h}^n = h^n(\bm{x}^n) + \bm{\epsilon}^n$ and $\bm{h}^{n-1} = h^{n-1}(\bm{x}^{n-1}) + \bm{\epsilon}^{n-1}$ with $\bm{\epsilon}^n \sim \mathcal{N}(0, \sigma^2 I)$ and $\bm{\epsilon}^{n-1} \sim \mathcal{N}(0, \sigma^2 I)$ being independent.
The negative log marginal likelihood (\ref{eq:NLML}) does not simply favor the models that fit the training data best. In fact, it induces an automatic trade-off between data-fit and model complexity. Specifically, minimizing the term $\bm{h}^{T}\bm{K}^{-1}\bm{h}$ in equation (\ref{eq:NLML}) targets fitting the training data, while the log-determinant term $\log |\bm{K}|$ penalizes model complexity. This regularization mechanism automatically meets the Occam's razor principle \cite{rasmussen2001occam} which encourages simplicity in explanations. The aforementioned regularization mechanism of the negative log marginal likelihood (\ref{eq:NLML}) effectively guards against overfitting and enables learning the unknown model parameters from very few\footnote{Regularization is important even in data abundant regimes as witnessed by the recently growing literature on discovering ordinary and partial differential equations from data using sparse regression techniques \cite{brunton2016discovering,Rudye1602614}.} noisy observations. However, there is no theoretical guarantee that the negative log marginal likelihood does not suffer from multiple local minima. Our practical experience so far with the negative log marginal likelihood seems to indicate that local minima are not a devastating problem, but certainly they do exist. Moreover, it should be highlighted that, although not pursued here, a fully Bayesian \cite{stuart2010inverse} and more robust estimate of the linear operator parameters $\lambda$ can be obtained by assigning priors on $\{\theta, \lambda, \sigma^2\}$. However, this would require more costly sampling procedures such as Markov Chain Monte Carlo (see \cite{Rasmussen06gaussianprocesses}, chapter 5) to train the model. Furthermore, the most computationally intensive part of learning using the negative log marginal likelihood (\ref{eq:NLML}) is associated with inverting dense covariance matrices $\bm{K}$. This scales cubically with the number $N$ of training data in $\bm{h}$. While it has been effectively addressed by the recent works of \cite{snelson2006sparse, hensman2013gaussian, raissi2017parametric}, this cubic scaling is still a well-known limitation of Gaussian process regression.

\section{Results}
The proposed framework provides a general treatment of time-dependent and nonlinear partial differential equations, which can be of fundamentally different nature. This generality will be demonstrated by applying the algorithm to a dataset originally proposed in \cite{Rudye1602614}, where sparse regression techniques are used to discover partial differential equations from time series measurements in the spatial domain. This dataset covers a wide range of canonical problems spanning a number of scientific domains including the Navier-Stokes, Schr\"{o}dinger, and Kuramoto-Sivashinsky equations. Moreover, all data and codes used in this manuscript are publicly available on GitHub at \url{https://github.com/maziarraissi/HPM}.

\subsection{Burgers' Equation}
\begin{figure}[!ht]
\centering
\includegraphics[width=\textwidth]{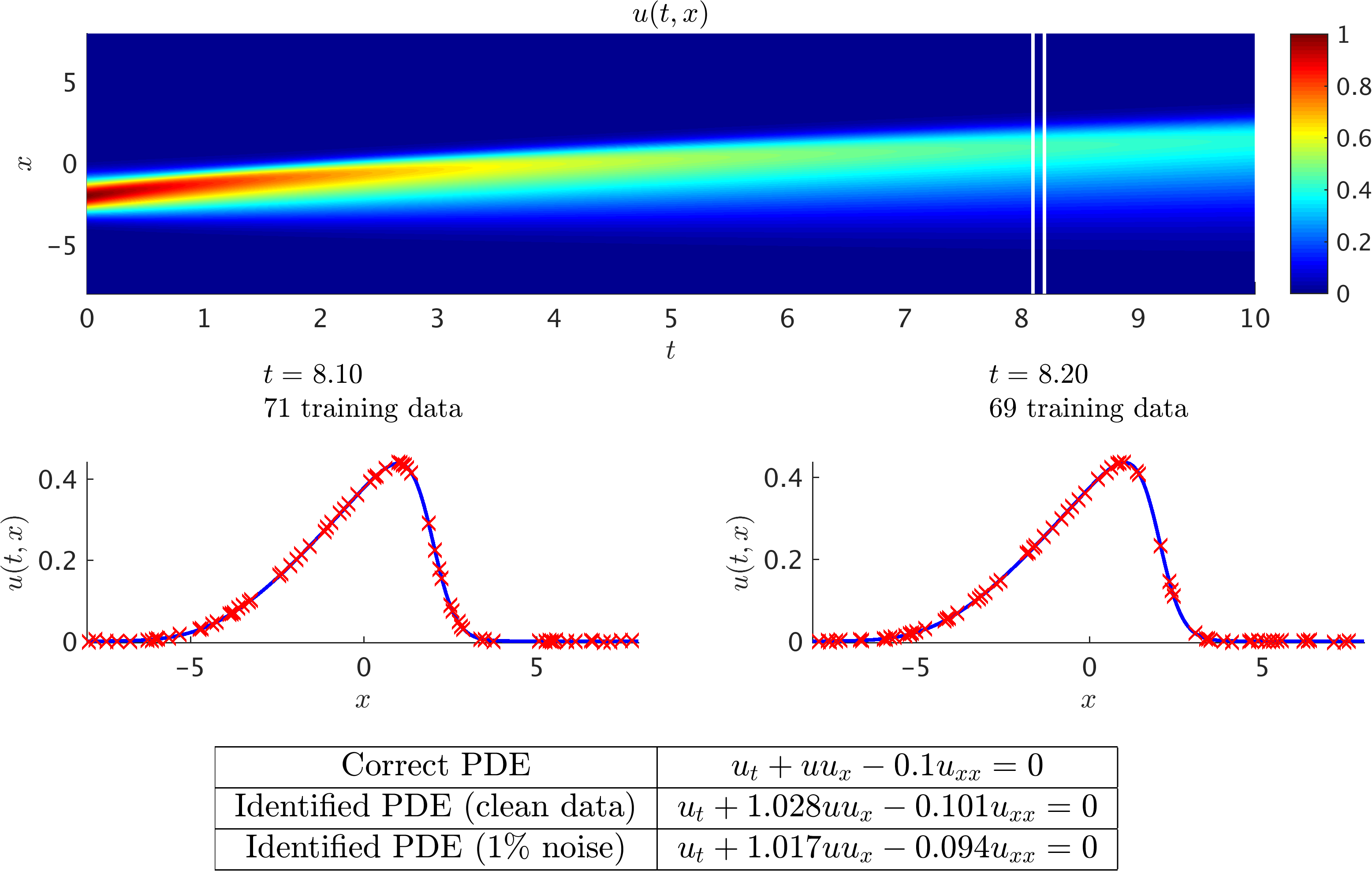}
\caption{\emph{Burgers' equation:} A solution to the Burgers' equation is depicted in the top panel. The two white vertical lines in this panel specify the locations of the two randomly selected snapshots. These two snapshots are $\Delta t = 0.1$ apart and are plotted in the middle panel. The red crosses denote the locations of the training data points. The correct partial differential equation along with the identified ones are reported in the lower panel.}\label{fig:Burgers}
\end{figure}

Burgers' equation arises in various areas of applied mathematics, including fluid mechanics, nonlinear acoustics, gas dynamics, and traffic flow \cite{basdevant1986spectral}. It is a fundamental partial differential equation and can be derived from the Navier-Stokes equations for the velocity field by dropping the pressure gradient term. Burgers' equation, despite its relation to the much more complicated Navier-Stokes equations, does not exhibit turbulent behavior. However, for small values of the viscosity parameters, Burgers' equation can lead to shock formation that is notoriously hard to resolve by classical numerical methods. In one space dimension the equation reads as
\begin{equation}\label{eq:Burgers}
u_t + \lambda_1 u u_x - \lambda_2 u_{xx} = 0,
\end{equation}
with $(\lambda_1, \lambda_2)$ being the unknown parameters. The original data-set proposed in \cite{Rudye1602614} contains 101 time snapshots of a solution to the Burgers' equation with a Gaussian initial condition, propagating into a traveling wave. The snapshots are $\Delta t = 0.1$ apart. The spatial discretization of each snapshot involves a uniform grid with 256 cells. As depicted in figure \ref{fig:Burgers} using {\em only two of these snapshots} (randomly selected) with 71 and 69 data points, respectively, the algorithm is capable of identifying the correct parameter values up to a relatively good accuracy. It should be noted that we are using only $140 = 71 + 69$ data points out of a total of $25856 = 101 \times 256$ in the original data set. This surprising performance is achieved at the cost of explicitly encoding the underlying physical laws expressed by the Burgers' equation in the covariance functions of the \emph{hidden physics model} (\ref{eq:HPM}). For a systematic study of the performance of the method, let us carry out the same experiment as the one illustrated in figure \ref{fig:Burgers} for every pair of consecutive snapshots in the original dataset. We are still using the same number of data points (i.e., 71 and 69) for each pair of snapshots, albeit in different locations. The resulting statistics for the learned parameter values are reported in table \ref{tab:Burgers1}. As is clearly demonstrated in this table, more noise in the data leads to less confidence in the estimated values for the parameters. Moreover, let us recall the main assumption of this work that the gap $\Delta t$ between the pair of snapshots should be small enough so that we can employ the backward Euler scheme (see equation (\ref{eq:BackwardEuler})). To test the importance of this assumption, let us use the exact same setup as the one explained in figure \ref{fig:Burgers}, but increase $\Delta t$. The results are reported in table \ref{tab:Burgers2}. Therefore, the most important facts about the proposed methodology are that more data, less noise, and a smaller gap $\Delta t$ between the two snapshots enhance the performance of the algorithm.
\begin{table}[!ht]
\begin{center}
\footnotesize
\begin{tabular}{|c|c|c|c|c|c|c|}
\hline
                & \multicolumn{2}{c|}{Clean Data} & \multicolumn{2}{c|}{$1\%$ Noise} & \multicolumn{2}{c|}{$5\%$ Noise} \\
\hline
                & $\lambda_1$ & $\lambda_2$ & $\lambda_1$ & $\lambda_2$ & $\lambda_1$ & $\lambda_2$ \\
\hline
First Quartile  & 1.0247 & 0.0942 & 0.9168 & 0.0784 & 0.3135 & 0.0027\\
\hline
Median          & 1.0379 & 0.0976 & 1.0274 & 0.0919 & 0.8294 & 0.0981\\
\hline
Third Quartile & 1.0555 & 0.0987 & 1.1161 & 0.1166 & 1.2488 & 0.1543\\
\hline
\end{tabular}
\end{center}
\caption{\emph{Burgers' equation:} Resulting statistics for the learned parameter values.}\label{tab:Burgers1}
\end{table}
\begin{table}[!ht]
\begin{center}
\footnotesize
\begin{tabular}{|c|c|c|c|c|c|c|}
\hline
                             &              & $\Delta t = 0.1$ & $\Delta t = 0.5$ & $\Delta t = 1.0$ & $\Delta t = 1.5$ \\
\hline
\multirow{2}{*}{Clean Data}  & $\lambda_1$  & 1.0283 & 1.1438 & 1.2500 & 1.2960\\
\cline{2-6}
                             & $\lambda_2$  & 0.1009 & 0.0934 & 0.0694 & 0.0431\\
\hline
\multirow{2}{*}{$1\%$ Noise} & $\lambda_1$  & 1.0170 & 1.1470 & 1.2584 & 1.3063\\
\cline{2-6}
                             & $\lambda_2$  & 0.0935 & 0.0939 & 0.0711 & 0.0428\\
\hline
\end{tabular}
\end{center}
\caption{\emph{Burgers' equation:} Effect of increasing the gap $\Delta t$ between the pair of snapshots.}\label{tab:Burgers2}
\end{table}

\subsection{The KdV Equation}
\begin{figure}[!ht]
\centering
\includegraphics[width=\textwidth]{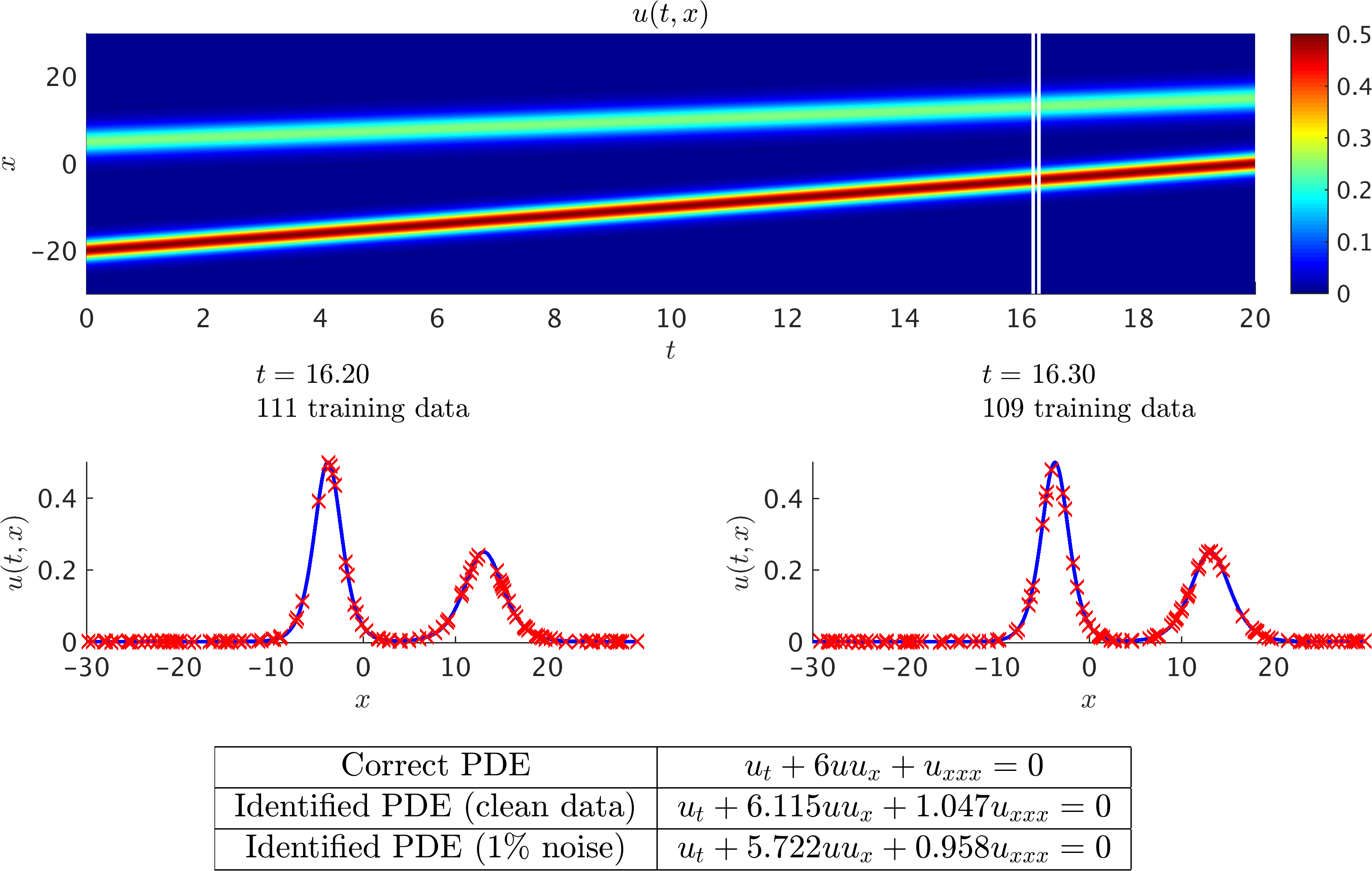}
\caption{\emph{The KdV equation:} A solution to the KdV equation is depicted in the top panel. The two white vertical lines in this panel specify the locations of the two randomly selected snapshots. These two snapshots are $\Delta t = 0.1$ apart and are plotted in the middle panel. The red crosses denote the locations of the training data points. The correct partial differential equation along with the identified ones are reported in the lower panel.}\label{fig:KDV}
\end{figure}

As a mathematical model of waves on shallow water surfaces one could consider the Korteweg-de Vries (KdV) equation. This equation can also be viewed as Burgers' equation with an added dispersive term. The KdV equation has several connections to physical problems. It describes the evolution of long one-dimensional waves in many physical settings. Such physical settings include shallow-water waves with weakly non-linear restoring forces, long internal waves in a density-stratified ocean, ion acoustic waves in a plasma, and acoustic waves on a crystal lattice. Moreover, the KdV equation is the governing equation of the string in the Fermi-Pasta-Ulam problem \cite{dauxois2008fermi} in the continuum limit. The KdV equation reads as
\begin{equation}\label{eq:KdV}
u_t + \lambda_1 u u_x + \lambda_2 u_{xxx} = 0,
\end{equation}
with $(\lambda_1, \lambda_2)$ being the unknown parameters. The original dataset proposed in \cite{Rudye1602614} contains a two soliton solution to the KdV equation with 512 spatial points and 201 time-steps. The snapshots are $\Delta t = 0.1$ apart. As depicted in figure \ref{fig:KDV} using only two of these snapshots (randomly selected) with 111 and 109 data points, respectively, the algorithm is capable of identifying the correct parameter values up to a relatively good accuracy. In particular, we are using $220 = 111 + 109$ out of a total of $102912 = 201 \times 512$ data points in the original data set. This level of efficiency is a direct consequence of equation (\ref{eq:HPM}) where the covariance functions explicitly encode the underlying physical laws expressed by the KdV equation. As a sensitivity analysis of the reported results, let us perform the same experiment as the one illustrated in figure \ref{fig:KDV} for every pair of consecutive snapshots in the original dataset. We are still using the same number of data points (i.e., 111 and 109) for each pair of snapshots, albeit in different locations. The resulting statistics for the learned parameter values are reported in table \ref{tab:KDV1}. As is clearly demonstrated in this table, more noise in the data leads to less confidence in the estimated values for the parameters. Moreover, to test the sensitivity of the results with respect to the gap between the two time snapshots, let us use the exact same setup as the one explained in figure \ref{fig:KDV}, but increase $\Delta t$. The results are reported in table \ref{tab:KDV2}. These results verify the most important facts about the proposed methodology that more data, less noise, and a smaller gap $\Delta t$ between the two snapshots enhance the performance of the algorithm.
\begin{table}[!ht]
\begin{center}
\footnotesize
\begin{tabular}{|c|c|c|c|c|c|c|}
\hline
                & \multicolumn{2}{c|}{Clean Data} & \multicolumn{2}{c|}{$1\%$ Noise} & \multicolumn{2}{c|}{$5\%$ Noise} \\
\hline
                & $\lambda_1$ & $\lambda_2$ & $\lambda_1$ & $\lambda_2$ & $\lambda_1$ & $\lambda_2$ \\
\hline
First Quartile  & 5.7783 & 0.9299 & 5.3358 & 0.7885 & 3.7435 & 0.2280\\
\hline
Median          & 5.8920 & 0.9656 & 5.5757 & 0.8777 & 4.5911 & 0.6060\\
\hline
Third Quartile  & 6.0358 & 1.0083 & 5.7840 & 0.9491 & 5.5106 & 0.8407\\
\hline
\end{tabular}
\end{center}
\caption{\emph{The KdV equation:} Resulting statistics for the learned parameter values.}\label{tab:KDV1}
\end{table}
\begin{table}[!ht]
\begin{center}
\footnotesize
\begin{tabular}{|c|c|c|c|c|c|c|c|}
\hline
                             &              & $\Delta t = 0.1$ & $\Delta t = 0.2$ & $\Delta t = 0.3$ & $\Delta t = 0.4$ & $\Delta t = 0.5$\\
\hline
\multirow{2}{*}{Clean Data}  & $\lambda_1$  & 6.1145 & 5.8948 & 5.4014 & 4.1779 & 3.5058 \\
\cline{2-7}
                             & $\lambda_2$  & 1.0470 & 0.9943 & 0.8535 & 0.4475 & 0.1816 \\
\hline
\multirow{2}{*}{$1\%$ Noise} & $\lambda_1$  & 5.7224 & 5.8288 & 5.4054 & 4.1479 & 3.4747 \\
\cline{2-7}
                             & $\lambda_2$  & 0.9578 & 0.9801 & 0.8563 & 0.4351 & 0.1622 \\
\hline
\end{tabular}
\end{center}
\caption{\emph{The KdV equation:} Effect of increasing the gap $\Delta t$ between the pair of snapshots.}\label{tab:KDV2}
\end{table}

\subsection{Kuramoto-Sivashinsky Equation}
\begin{figure}[!ht]
\centering
\includegraphics[width=\textwidth]{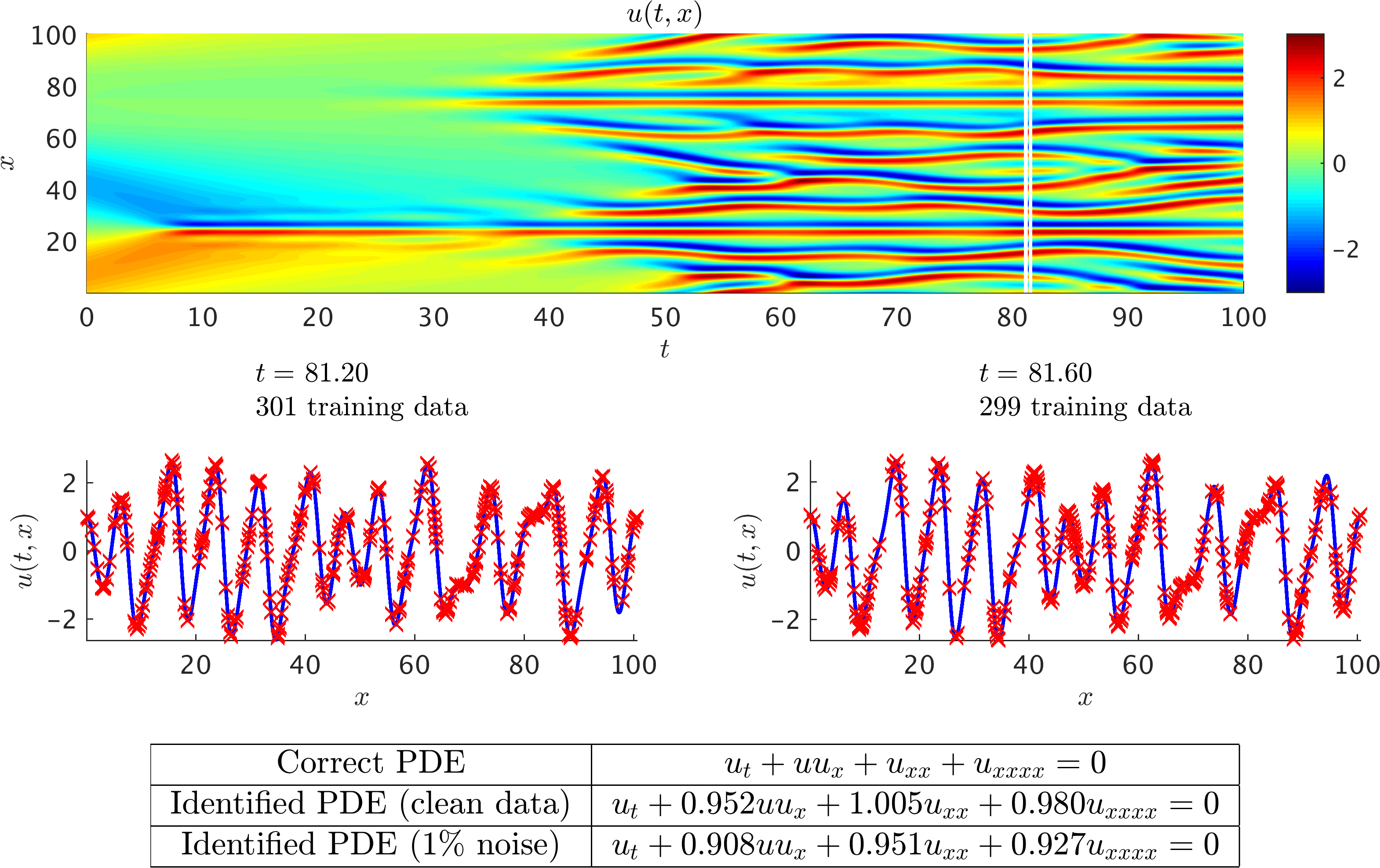}
\caption{\emph{Kuramoto-Sivashinsky equation:} A solution to the Kuramoto-Sivashinsky equation is depicted in the top panel. The two white vertical lines in this panel specify the locations of the two randomly selected snapshots. These two snapshots are $\Delta t = 0.4$ apart and are plotted in the middle panel. The red crosses denote the locations of the training data points. The correct partial differential equation along with the identified ones are reported in the lower panel.}\label{fig:KS}
\end{figure}

The Kuramoto-Sivashinsky equation \cite{hyman1986kuramoto, shraiman1986order, nicolaenko1985some} has similarities with Burgers' equation. However, because of the presence of both second and fourth order spatial derivatives, its behavior is far more complicated and interesting. The Kuramoto-Sivashinsky is a canonical model of a pattern forming system with spatio-temporal chaotic behavior. The sign of the second derivative term is such that it acts as an energy source and thus has a destabilizing effect. The nonlinear term, however, transfers energy from low to high wave numbers where the stabilizing fourth derivative term dominates. The first derivation of this equation was by Kuramoto in the study of reaction-diffusion equations modeling the Belousov-Zabotinskii reaction. The equation was also developed by Sivashinsky in higher space dimensions in modeling small thermal diffusive instabilities in laminar flame fronts and in small perturbations from a reference Poiseuille flow of a film layer on an inclined plane. In one space dimension it has also been used as a model for the problem of B\'{e}nard convection in an elongated box, and it may be used to describe long waves on the interface between two viscous fluids and unstable drift waves in plasmas. In one space dimension the Kuramoto-Sivashinsky equation reads as
\begin{equation}\label{eq:KS}
u_t + \lambda_1 u u_x + \lambda_2 u_{xx} + \lambda_3 u_{xxxx} = 0,
\end{equation}
where $(\lambda_1, \lambda_2, \lambda_3)$ are the unknown parameters. The original dataset proposed in \cite{Rudye1602614} contains a direct numerical solution of the Kuramoto-Sivashinsky equation with 1024 spatial points and 251 time-steps. The snapshots are $\Delta t = 0.4$ apart. As depicted in figure \ref{fig:KS} using only two of these snapshots (randomly selected) with 301 and 299 data points, respectively, the algorithm is capable of identifying the correct parameter values up to a relatively good accuracy. In particular, we are using $600 = 301 + 299$ out of a total of $257024 = 251 \times 1024$ data points in the original data set. This is possible because of equation (\ref{eq:HPM}) where the covariance functions explicitly encode the underlying physical laws expressed by the Kuramoto-Sivashinsky equation. For a sensitivity analysis of the reported results, let us perform the same experiment as the one illustrated in figure \ref{fig:KS} for every pair of consecutive snapshots in the original dataset. We are still using the same number of data points (i.e., 301 and 299) for each pair of snapshots, albeit in different locations. The resulting statistics for the learned parameter values are reported in table \ref{tab:KS1}. As shown in this table, more noise in the data leads to less confidence in the estimated parameter values. Moreover, to test the sensitivity of the results with respect to the gap between the two time snapshots, let us use the exact same setup as the one explained in figure \ref{fig:KS}, but increase $\Delta t$. The results are reported in table \ref{tab:KS2}. These results indicate that more data, less noise, and a smaller gap $\Delta t$ between the two snapshots enhance the performance of the algorithm.
\begin{table}[!ht]
\begin{center}
\scriptsize
\begin{tabular}{|c|c|c|c|c|c|c|c|c|c|}
\hline
                & \multicolumn{3}{c|}{Clean Data} & \multicolumn{3}{c|}{$1\%$ Noise} & \multicolumn{3}{c|}{$5\%$ Noise} \\
\hline
                & $\lambda_1$ & $\lambda_2$ & $\lambda_3$ & $\lambda_1$ & $\lambda_2$ & $\lambda_3$ & $\lambda_1$ & $\lambda_2$ & $\lambda_3$ \\
\hline
First Quartile  & 0.9603 & 0.9829 & 0.9711 & 0.7871 & 0.8095 & 0.5891 & -0.0768 & 0.0834 & -0.0887\\
\hline
Median          & 0.9885 & 1.0157 & 0.9970 & 0.8746 & 0.9124 & 0.8798 & 0.4758 & 0.5539 & 0.4086\\
\hline
Third Quartile  & 1.0187 & 1.0550 & 1.0314 & 0.9565 & 0.9948 & 0.9553 & 0.6991 & 0.7644 & 0.7009\\
\hline
\end{tabular}
\end{center}
\caption{\emph{Kuramoto-Sivashinsky equation:} Resulting statistics for the learned parameter values.}\label{tab:KS1}
\end{table}
\begin{table}[!ht]
\begin{center}
\footnotesize
\begin{tabular}{|c|c|c|c|c|c|}
\hline
                             &              & $\Delta t = 0.4$ & $\Delta t = 0.8$ & $\Delta t = 1.2$\\
\hline
\multirow{2}{*}{Clean Data}  & $\lambda_1$  & 0.9515 & 0.5299 & 0.1757\\
\cline{2-5}
                             & $\lambda_2$  & 1.0052 & 0.5614 & 0.1609\\
\cline{2-5}                             
                             & $\lambda_3$  & 0.9803 & 0.5438 & 0.1647\\
\hline
\multirow{3}{*}{$1\%$ Noise} & $\lambda_1$  & 0.9081 & 0.5124 & 0.1616\\
\cline{2-5}
                             & $\lambda_2$  & 0.9511 & 0.5387 & 0.1436\\
\cline{2-5}                             
                             & $\lambda_3$  & 0.9266 & 0.5213 & 0.1483\\
\hline
\end{tabular}
\end{center}
\caption{\emph{Kuramoto-Sivashinsky equation:} Effect of increasing the gap $\Delta t$ between the pair of snapshots.}\label{tab:KS2}
\end{table}

\subsection{Nonlinear Schr\"{o}dinger Equation}
\begin{figure}
\centering
\includegraphics[width=\textwidth]{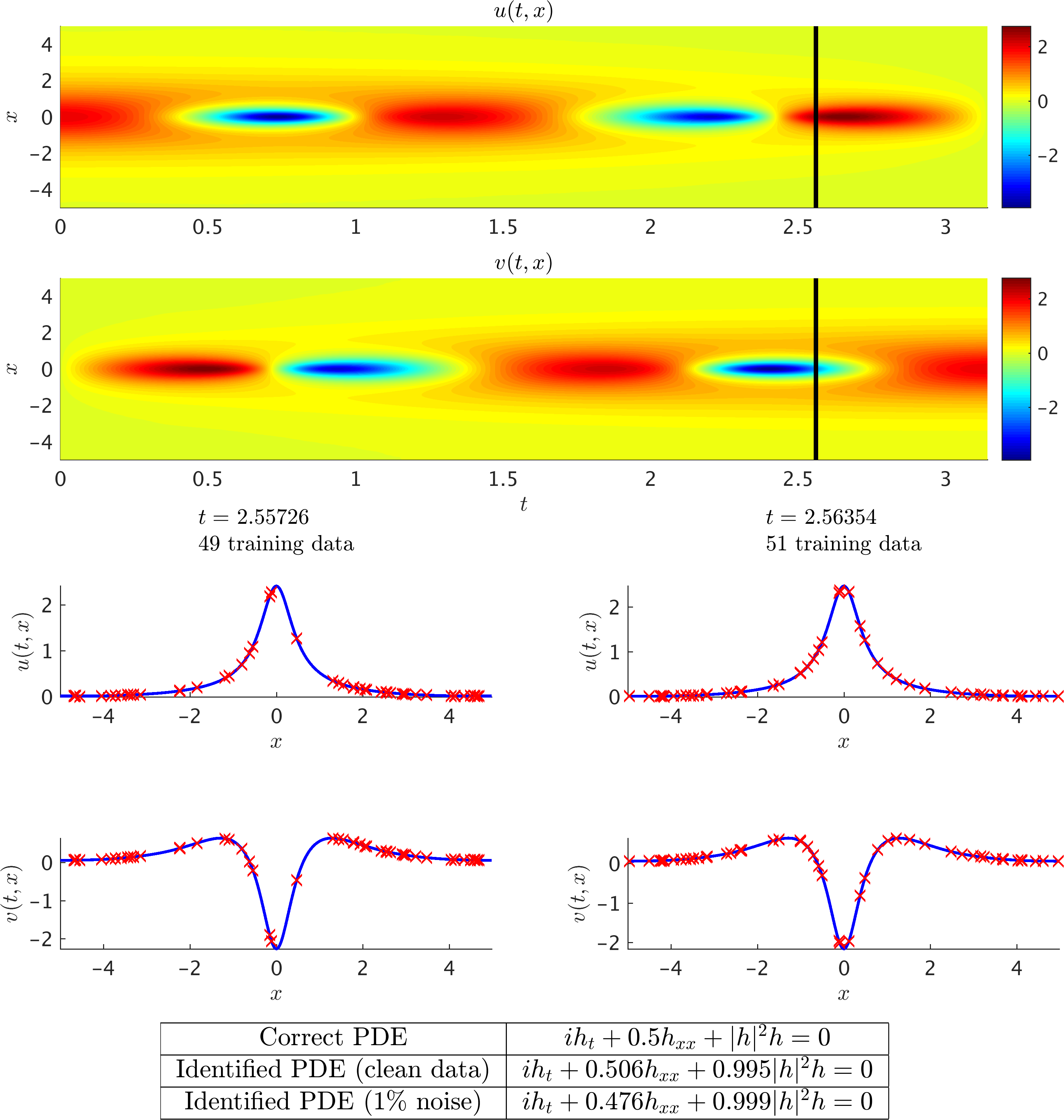}
\caption{\emph{Nonlinear Schr\"{o}dinger equation:} A solution to the nonlinear Schr\"{o}dinger equation is depicted in the top two panels. The two black vertical lines in these two panels specify the locations of the two randomly selected snapshots. These two snapshots are $\Delta t = 0.0063$ apart and are plotted in the two middle panels. The red crosses denote the locations of the training data points. The correct partial differential equation along with the identified ones are reported in the lower panel. Here, $u$ is the real part of $h$ and $v$ is the imaginary part.}\label{fig:Schrodinger}
\end{figure}

The one-dimensional nonlinear Schr\"{o}dinger equation is a classical field equation that is used to study nonlinear wave propagation in optical fibers and/or waveguides, Bose-Einstein condensates, and plasma waves. In optics, the nonlinear term arises from the intensity dependent index of refraction of a given material. Similarly, the nonlinear term for Bose-Einstein condensates is a result of the mean-field interactions of an interacting, N-body system. The nonlinear Schr\"{o}dinger equation is given by
\begin{equation}\label{eq:Schrodinger}
i h_t + \lambda_1 h_{xx} + \lambda_2 |h|^2 h = 0,
\end{equation}
where $(\lambda_1, \lambda_2)$ are the unknown parameters. Let $u$ denote the real part of $h$ and $v$ the imaginary part. Then, the nonlinear Schr\"{o}dinger equation can be equivalently written as
\begin{eqnarray}
u_t + \lambda_1 v_{xx} + \lambda_2 (u^2 + v^2) v = 0,\label{eq:SchrodingerSystem}\\
v_t - \lambda_1 u_{xx} - \lambda_2 (u^2 + v^2) u = 0.\nonumber
\end{eqnarray}
Employing the backward Euler time stepping scheme, we obtain
\begin{eqnarray}
u^n + \Delta t \lambda_1 v_{xx}^n + \Delta t \lambda_2 [(u^n)^2 + (v^n)^2] v^n = u^{n-1},\label{eq:SchrodingerSystemBackwardEuler}\\
v^n - \Delta t \lambda_1 u_{xx}^n - \Delta t \lambda_2 [(u^n)^2 + (v^n)^2] u^n = v^{n-1}.\nonumber
\end{eqnarray}
The above equations can be approximated by
\begin{eqnarray}
u^n + \Delta t \lambda_1 v_{xx}^n + \Delta t \lambda_2 [(u^{n-1})^2 + (v^{n-1})^2] v^n = u^{n-1},\label{eq:SchrodingerSystemBackwardEulerLinear}\\
v^n - \Delta t \lambda_1 u_{xx}^n - \Delta t \lambda_2 [(u^{n-1})^2 + (v^{n-1})^2] u^n = v^{n-1},\nonumber
\end{eqnarray}
which involves only linear operations. Here, $u^{n-1}(x)$ and $v^{n-1}(x)$ are the real and imaginary parts of the state of the system at the previous time step, respectively. We proceed by placing two independent Gaussian processes on $u^n(x)$ and $v^n(x)$; i.e.,
\begin{eqnarray}\label{eq:SchrodingerPriors}
&& u^n(x) \sim \mathcal{GP}(0, k_u(x,x';\theta_u)),\\
&& v^n(x) \sim \mathcal{GP}(0, k_v(x,x';\theta_v)).\nonumber
\end{eqnarray}
Here, $\theta_u$ and $\theta_v$ are the hyper-parameters of the kernels $k_u$ and $k_v$, respectively. The prior assumptions (\ref{eq:SchrodingerPriors}) along with equations (\ref{eq:SchrodingerSystemBackwardEulerLinear}) enable us to encode the underlying laws of physics expressed by the nonlinear Schr\"{o}dinger equation in the resulting \emph{hidden physics model}
\begin{equation}\label{eq:SchrodingerHPM}
\begin{bmatrix}
u^n\\
v^n\\
u^{n-1}\\
v^{n-1}
\end{bmatrix}
\sim
\mathcal{GP}\left(0, \begin{bmatrix}
k^{n,n}_{u,u} & k^{n,n}_{u,v} & k^{n,n-1}_{u,u} & k^{n,n-1}_{u,v}\\
k^{n,n}_{v,u} & k^{n,n}_{v,v} & k^{n,n-1}_{v,u} & k^{n,n-1}_{v,v}\\
k^{n-1,n}_{u,u} & k^{n-1,n}_{u,v} & k^{n-1,n-1}_{u,u} & k^{n-1,n-1}_{u,v}\\
k^{n-1,n}_{v,u} & k^{n-1,n}_{v,v} & k^{n-1,n-1}_{v,u} & k^{n-1,n-1}_{v,v}\\
\end{bmatrix}
\right).
\end{equation}
The specific forms of the covariance functions involved in model (\ref{eq:SchrodingerHPM}) is a direct function of the prior assumptions (\ref{eq:SchrodingerPriors}) as well as equations (\ref{eq:SchrodingerSystemBackwardEulerLinear}). The hyper-parameters $\theta_u$ and $\theta_v$ along with the parameters $\lambda_1$ and $\lambda_2$ are learned by minimizing the negative log marginal likelihood as outlined in section \ref{sec:Learning}. The original data-set proposed in \cite{Rudye1602614} contains 501 time snapshots of a solution to the nonlinear Schr\"{o}dinger equation with a Gaussian initial condition. The snapshots are $\Delta t = 0.0063$ apart. The spatial discretization of each snapshot involves a uniform grid with 512 elements. As depicted in figure \ref{fig:Schrodinger} using only two of these snapshots (randomly selected) with 49 and 51 data points, respectively, the algorithm is capable of identifying the correct parameter values up to a relatively good accuracy. It should be noted that we are using only $100 = 49 + 51$ data points out of a total of $256512 = 501 \times 512$ in the original data set. Such a performance is achieved at the cost of explicitly encoding the underlying physical laws expressed by the nonlinear Schr\"{o}dinger equation in the covariance functions of the \emph{hidden physics model} (\ref{eq:SchrodingerHPM}). For a systematic study of the performance of the method, let us carry out the same experiment as the one illustrated in figure \ref{fig:Schrodinger} for every pair of consecutive snapshots in the original dataset. We are still using the same number of data points (i.e., 49 and 51) for each pair of snapshots. The resulting statistics for the learned parameter values are reported in table \ref{tab:Schrodinger1}. As is clearly demonstrated in this table, more noise in the data leads to less confidence in the estimated values for the parameters. Moreover, let us recall the main assumption of this work that the gap $\Delta t$ between the pair of snapshots should be small enough so that we can employ the backward Euler scheme (see equation (\ref{eq:SchrodingerSystemBackwardEuler})). To test the importance of this assumption, let us use the exact same setup as the one explained in figure \ref{fig:Schrodinger}, but increase $\Delta t$. The results are reported in table \ref{tab:Schrodinger2}. Therefore, the most important facts about the proposed methodology are that more data, less noise, and a smaller gap $\Delta t$ between the two snapshots enhance the performance of the algorithm.
\begin{table}[!ht]
\begin{center}
\footnotesize
\begin{tabular}{|c|c|c|c|c|c|c|}
\hline
                & \multicolumn{2}{c|}{Clean Data} & \multicolumn{2}{c|}{$1\%$ Noise} & \multicolumn{2}{c|}{$5\%$ Noise} \\
\hline
                & $\lambda_1$ & $\lambda_2$ & $\lambda_1$ & $\lambda_2$ & $\lambda_1$ & $\lambda_2$ \\
\hline
First Quartile  & 0.4950 & 0.9960 & 0.3714 & 0.9250 & -0.1186 & 0.6993\\
\hline
Median          & 0.5009 & 1.0001 & 0.4713 & 0.9946 & 0.4259 & 0.9651\\
\hline
Third Quartile  & 0.5072 & 1.0039 & 0.5918 & 1.0670 & 0.9730 & 1.2730\\
\hline
\end{tabular}
\end{center}
\caption{\emph{Nonlinear Schr\"{o}dinger equation:} Resulting statistics for the learned parameter values.}\label{tab:Schrodinger1}
\end{table}
\begin{table}[!ht]
\begin{center}
\footnotesize
\begin{tabular}{|c|c|c|c|c|c|c|}
\hline
                             &              & $\Delta t = 0.0063$ & $\Delta t = 0.0628$ & $\Delta t = 0.1257$ & $\Delta t = 0.1885$ \\
\hline
\multirow{2}{*}{Clean Data}  & $\lambda_1$  & 0.5062 & 0.4981 & 0.3887 & 0.3097\\
\cline{2-6}
                             & $\lambda_2$  & 0.9949 & 0.8987 & 0.7936 & 0.7221\\
\hline
\multirow{2}{*}{$1\%$ Noise} & $\lambda_1$  & 0.4758 & 0.4976 & 0.3928 & 0.3128\\
\cline{2-6}
                             & $\lambda_2$  & 0.9992 & 0.9011 & 0.7975 & 0.7255\\
\hline
\end{tabular}
\end{center}
\caption{\emph{Nonlinear Schr\"{o}dinger equation:} Effect of increasing the gap $\Delta t$ between the pair of snapshots.}\label{tab:Schrodinger2}
\end{table}

\subsection{Navier-Stokes Equations}
\begin{figure}
\centering
\includegraphics[width=\textwidth]{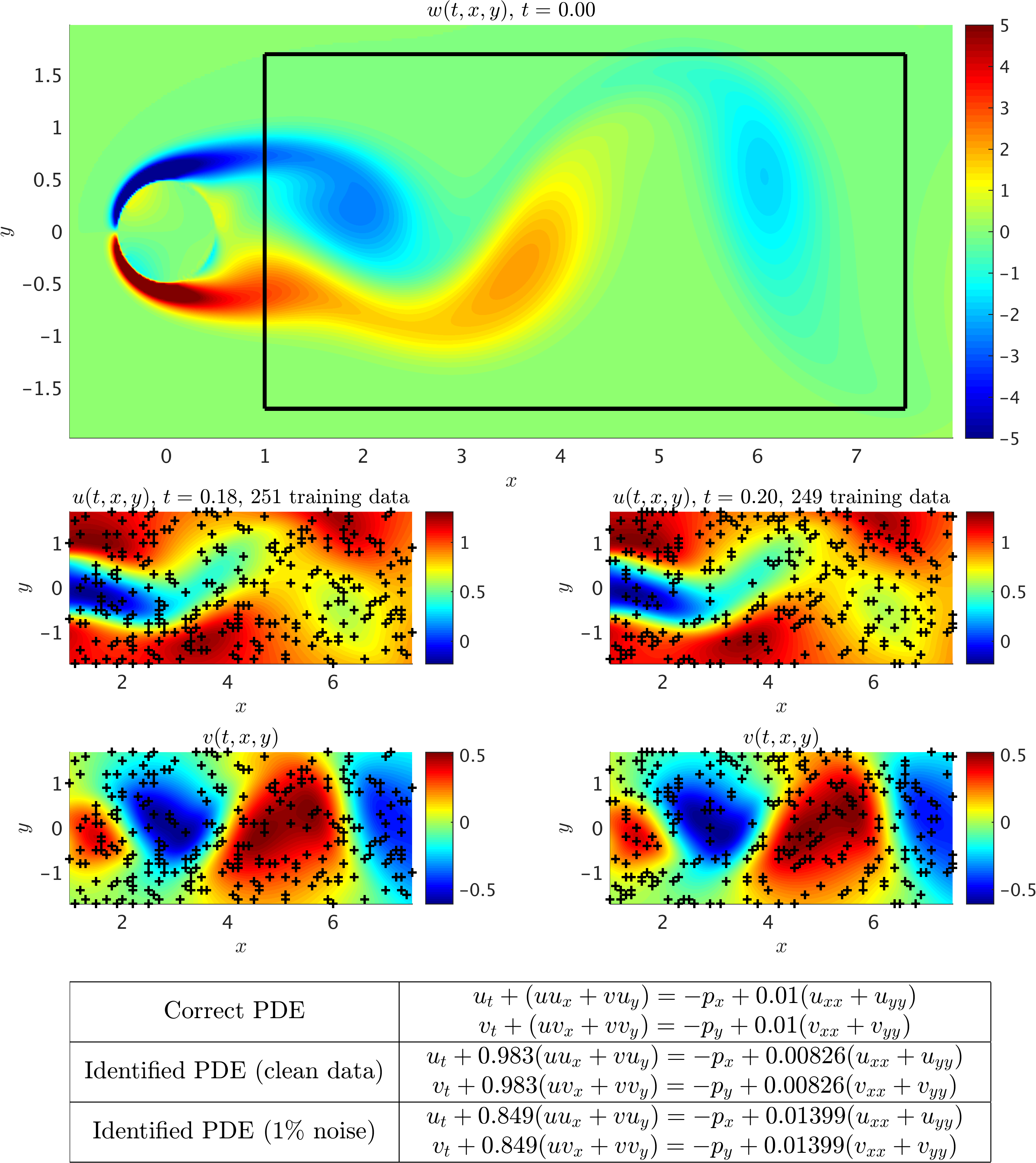}
\caption{\emph{Navier-Stokes equations:} A single snapshot of the vorticity field of a solution to the Navier-Stokes equations for the fluid flow past a cylinder is depicted in the top panel. The black box in this panel specifies the sampling region. Two snapshots of the velocity field being $\Delta t = 0.02$ apart are plotted in the two middle panels. The black crosses denote the locations of the training data points. The correct partial differential equation along with the identified ones are reported in the lower panel. Here, $u$ denotes the $x$-component of the velocity field, $v$ the $y$-component, $p$ the pressure, and $w$ the vorticity field.}\label{fig:NavierStokes}
\end{figure}

Navier-Stokes equations describe the physics of many phenomena of scientific and engineering interest. They may be used to model the weather, ocean currents, water flow in a pipe and air flow around a wing. The Navier-Stokes equations in their full and simplified forms help with the design of aircraft and cars, the study of blood flow, the design of power stations, the analysis of the dispersion of pollutants, and many other applications. Let us consider the Navier-Stokes equations in two dimensions\footnote{It is straightforward to generalize the proposed framework to the Navier-Stokes equations in three dimensions (3D).} (2D) given explicitly by
\begin{equation}\label{eq:NavierStokes}
\begin{array}{c}
u_t + \lambda_1 (u u_x + v u_y) = -p_x + \lambda_2(u_{xx} + u_{yy}),\\
v_t + \lambda_1 (u v_x + v v_y) = -p_y + \lambda_2(v_{xx} + v_{yy}),
\end{array}
\end{equation}
where $u(t, x, y)$ denotes the $x$-component of the velocity field, $v(t, x, y)$ the $y$-component, and $p(t, x, y)$ the pressure. Here, $\lambda = (\lambda_1, \lambda_2)$ are the unknown parameters. Solutions to the Navier-Stokes equations are searched in the set of divergence-free functions; i.e.,
\begin{equation}\label{eq:Continuity}
u_x + v_y = 0.
\end{equation}
This extra equation is the continuity equation for incompressible fluids that describes the conservation of mass of the fluid. Applying the backward Euler time stepping scheme to the Navier-Stokes equations (\ref{eq:NavierStokes}) we obtain
\begin{equation}\label{eq:NavierStokesBackwardEuler}
\begin{array}{c}
u^n + \Delta t \lambda_1 (u^n u_x^n + v^n u_y^n) + \Delta t p_x^n - \Delta t \lambda_2 (u_{xx}^n + u_{yy}^n) = u^{n-1},\\
v^n + \Delta t \lambda_1 (u^n v_x^n + v^n v_y^n) + \Delta t p_y^n - \Delta t \lambda_2 (v_{xx}^n + v_{yy}^n) = v^{n-1},
\end{array}
\end{equation}
where $u^n(x,y) = u(t^n, x, y)$ and $v^n(x,y) = v(t^n, x, y)$. We make the assumption that
\begin{equation}\label{eq:streamline}
u^n = \psi^n_y,\ \ \ v^n = -\psi^n_x,
\end{equation}
for some latent function $\psi^n(x,y)$. Under this assumption, the continuity equation (\ref{eq:Continuity}) will be automatically satisfied. We proceed by placing a Gaussian process prior on
\begin{equation}
\psi^n(x,y) \sim \mathcal{GP}\left(0, k((x,y), (x',y');\theta)\right),
\end{equation}
where $\theta$ are the hyper-parameters of the kernel $k((x,y), (x',y');\theta)$. This will result in the following multi-output Gaussian process
\begin{equation}\label{eq:ContinuityGP}
\begin{bmatrix}
u^n\\
v^n
\end{bmatrix} \sim \mathcal{GP}\left(0, \begin{bmatrix}
k_{u,u}^{n,n} & k_{u,v}^{n,n} \\
k_{v,u}^{n,n} & k_{v,v}^{n,n}
\end{bmatrix}\right),
\end{equation}
where
\begin{align*}
&k^{n,n}_{u,u} = \frac{\partial}{\partial y} \frac{\partial}{\partial y'} k, && k^{n,n}_{u,v} = -\frac{\partial}{\partial y} \frac{\partial}{\partial x'} k,\\
&k^{n,n}_{v,u} = -\frac{\partial}{\partial x} \frac{\partial}{\partial y'} k, &&  k^{n,n}_{v,v} = \frac{\partial}{\partial x} \frac{\partial}{\partial x'} k.\nonumber
\end{align*}
By construction (see equation (\ref{eq:streamline})), any samples generated from this multi-output Gaussian process will satisfy the continuity equation (\ref{eq:Continuity}). Moreover, independent from $\psi^n(x, y)$, we will place a Gaussian process prior on $p^n(x, y)$; i.e.,
\begin{eqnarray}
p^n(x, y) \sim \mathcal{GP}(0, k_{p,p}^{n,n}( (x,y), (x',y');\theta_p)).
\end{eqnarray}
We linearize the backward Euler time stepping scheme by employing the states $u^{n-1}(x,y)$ and $v^{n-1}(x,y)$ of the system at the previous time step and writing
\begin{equation}\label{eq:NavierStokesBackwardEulerLinearized}
\begin{array}{c}
u^n + \Delta t \lambda_1 (u^{n-1} u_x^n + v^{n-1} u_y^n) + \Delta t p_x^n - \Delta t \lambda_2(u_{xx}^n + u_{yy}^n) = u^{n-1},\\
v^n + \Delta t \lambda_2 (u^{n-1} v_x^n + v^{n-1} v_y^n) + \Delta t p_y^n - \Delta t \lambda_2(v_{xx}^n + v_{yy}^n) = v^{n-1}.
\end{array}
\end{equation}
The above equations (\ref{eq:NavierStokesBackwardEulerLinearized}) can be rewritten as
\begin{equation}
\begin{array}{c}
\mathcal{L}_{(x,y)}^\lambda u^n + \Delta t p_x^n = u^{n-1},\\
\mathcal{L}_{(x,y)}^\lambda v^n + \Delta t p_y^n = v^{n-1},
\end{array}
\end{equation}
by defining the linear operator $\mathcal{L}_{(x,y)}^\lambda$ to be given by
\begin{equation}
\mathcal{L}_{(x,y)}^\lambda h := h + \Delta t \lambda_1 (u^{n-1} h_x + v^{n-1} h_y) - \Delta t \lambda_2 (h_{xx} + h_{yy}).
\end{equation}
This will allow us to obtain the following \emph{hidden physics model} encoding the structure of the Navier-Stokes equations and the backward Euler time stepping scheme in its kernels; i.e.,
\begin{equation}\label{eq:NavierStokesHPM}
\begin{bmatrix}
u^n\\
v^n\\
p^n\\
u^{n-1}\\
v^{n-1}\\
\end{bmatrix} \sim \mathcal{GP}\left(0, \begin{bmatrix}
k_{u,u}^{n,n} & k_{u,v}^{n,n} & 0 & k_{u,u}^{n,n-1} & k_{u,v}^{n,n-1}\\
              & k_{v,v}^{n,n} & 0 & k_{v,u}^{n,n-1} & k_{v,v}^{n,n-1}\\
              &   & k_{p,p}^{n,n} & k_{p,u}^{n,n-1} & k_{p,v}^{n,n-1}\\
              &   &             & k_{u,u}^{n-1,n-1} & k_{u,v}^{n-1,n-1}\\
              &   &             &                   & k_{v,v}^{n-1,n-1}\\
\end{bmatrix}\right),
\end{equation}
where 
\begin{eqnarray*}
k^{n,n-1}_{u,u} = \mathcal{L}_{(x',y')}^\lambda k^{n,n}_{u,u}, && k^{n,n-1}_{u,v} = \mathcal{L}_{(x',y')}^\lambda k^{n,n}_{u,v},\\
k^{n,n-1}_{v,u} = \mathcal{L}_{(x',y')}^\lambda k^{n,n}_{v,u}, &&  k^{n,n-1}_{v,v} = \mathcal{L}_{(x',y')}^\lambda k^{n,n}_{v,v},\\
k^{n,n-1}_{p,u} = \Delta t \frac{\partial }{\partial x'} k_{p,p}^{n,n}, &&  k^{n,n-1}_{p,v} = \Delta t \frac{\partial }{\partial y'} k_{p,p}^{n,n},
\end{eqnarray*}
and
\begin{eqnarray*}
&&k^{n-1,n-1}_{u,u} = \mathcal{L}_{(x,y)}^\lambda k^{n,n-1}_{u,u} + \Delta t \frac{\partial }{\partial x}k^{n,n-1}_{p,u},\\ 
&&k^{n-1,n-1}_{u,v} = \mathcal{L}_{(x,y)}^\lambda k^{n,n-1}_{u,v} + \Delta t \frac{\partial }{\partial x}k^{n,n-1}_{p,v},\\
&&k^{n-1,n-1}_{v,v} = \mathcal{L}_{(x,y)}^\lambda k^{n,n-1}_{v,v} + \Delta t \frac{\partial }{\partial y}k^{n,n-1}_{p,v}.
\end{eqnarray*}
The lower triangular portion of the matrix of covariance functions (\ref{eq:NavierStokesHPM}) is not shown due to symmetry. The hyper-parameters $\theta$ and $\theta_p$ along with the parameters $\lambda = (\lambda_1, \lambda_2)$ are learned by minimizing the negative log marginal likelihood as outlined in section \ref{sec:Learning}. As for the data, following the exact same instructions as the ones provided in \cite{kutz2016dynamic} and \cite{Rudye1602614}, we simulate the Navier-Stokes equations describing the two-dimensional fluid flow past a circular cylinder at Reynolds number 100 using the Immersed Boundary Projection Method \cite{taira2007immersed, colonius2008fast}. This approach utilizes a multi-domain scheme with four nested domains, each successive grid being twice as large as the previous one. Length and time are nondimensionalized so that the cylinder has unit diameter and the flow has unit velocity. Data is collected on the finest domain with dimensions $9 \times 4$ at a grid resolution of $449 \times 199$. The flow solver uses a 3rd-order Runge Kutta integration scheme with a time step of t = 0.02, which has been verified to yield well-resolved and converged flow fields. After simulations converge to steady periodic vortex shedding, flow snapshots are saved every $\Delta t = 0.02$. As depicted in figure \ref{fig:NavierStokes} using only two snapshots of the velocity\footnote{It is worth emphasizing that we are not making use of any data on the pressure or vorticity fields. In practice, unlike velocity (e.g., Particle Image Velocimetry (PIV) data), obtaining direct measurements of the pressure or vorticity fields are more demanding if not impossible. Our method circumvents the need for having data on the pressure simply because of the prior assumption (\ref{eq:ContinuityGP}) where any samples generated from this multi-output Gaussian process satisfy the continuity equation (\ref{eq:Continuity}). } field with 251 and 249 data points, respectively, the algorithm is capable of identifying the correct parameter values up to a relatively good accuracy. It should be noted that we are using only two snapshots with a total of $500 = 251 + 249$ data points. This surprising performance is achieved at the cost of explicitly encoding the underlying physical laws expressed by the Navier-Stokes equations in the covariance functions of the \emph{hidden physics model} (\ref{eq:NavierStokesHPM}). For a sensitivity analysis of the reported results, let us perform the same experiment as the one illustrated in figure \ref{fig:NavierStokes} for 501 pairs of consecutive snapshots. We are still using the same number of data points (i.e., 251 and 249) for each pair of snapshots. The resulting statistics for the learned parameter values are reported in table \ref{tab:NavierStokes1}. As is clearly demonstrated in this table, more noise in the data leads to less confidence in the estimated values for the parameters. Moreover, to test the sensitivity of the results with respect to the gap between two time snapshots, let us use the exact same setup as the one explained in figure \ref{fig:NavierStokes}, but increase $\Delta t$. The results are reported in table \ref{tab:NavierStokes2}. These results verify the most important facts about the proposed methodology that more data, less noise, and a smaller gap $\Delta t$ between the two snapshots enhance the performance of the algorithm. In particular, the results reported in table \ref{tab:NavierStokes2} indicate that to obtain more accurate estimates of the Reynolds number $1/\lambda_2$ one needs to utilize a smaller gap $\Delta t$ between the pair of snapshots.  To verify the validity of this conjecture let us decrease the gap $\Delta t$ between the pair of time snapshots while employing the exact same setup as the one explained in figure \ref{fig:NavierStokes}. The results are reported in table \ref{tab:NavierStokes3}. As is clearly demonstrated in this table, a smaller $\Delta t$ leads to more accurate estimates of the Reynolds number $1/\lambda_2$ in the absence of noise in the data. However, a smaller $\Delta t$ seems to make the algorithm more susceptible to noise in the data.
\begin{table}[!ht]
\begin{center}
\footnotesize
\begin{tabular}{|c|c|c|c|c|c|c|}
\hline
                & \multicolumn{2}{c|}{Clean Data} & \multicolumn{2}{c|}{$1\%$ Noise} & \multicolumn{2}{c|}{$5\%$ Noise} \\
\hline
                & $\lambda_1$ & $\lambda_2$ & $\lambda_1$ & $\lambda_2$ & $\lambda_1$ & $\lambda_2$ \\
\hline
First Quartile  & 0.9854 & 0.0069 & 0.8323 & 0.0057 & 0.5373 & 0.0026\\
\hline
Median          & 0.9928 & 0.0077 & 0.8717 & 0.0063 & 0.6498 & 0.0030\\
\hline
Third Quartile  & 1.0001 & 0.0086 & 0.9102 & 0.0070 & 0.7619 & 0.0046\\
\hline
\end{tabular}
\end{center}
\caption{\emph{Navier-Stokes equations:} Resulting statistics for the learned parameter values.}\label{tab:NavierStokes1}
\end{table}
\begin{table}[!ht]
\begin{center}
\footnotesize
\begin{tabular}{|c|c|c|c|c|c|c|c|}
\hline
                             &              & $\Delta t = 0.02$ & $\Delta t = 0.04$ & $\Delta t = 0.06$ & $\Delta t = 0.08$ & $\Delta t = 1.0$\\
\hline
\multirow{2}{*}{Clean Data}  & $\lambda_1$  & 0.9834 & 0.9925 & 0.9955 & 0.9976 & 1.0021 \\
\cline{2-7}
                             & $\lambda_2$  & 0.0083 & 0.0072 & 0.0058 & 0.0040 & 0.0027 \\
\hline
\multirow{2}{*}{$1\%$ Noise} & $\lambda_1$  & 0.8488 & 0.9298 & 0.9597 & 0.9726 & 0.9791 \\
\cline{2-7}
                             & $\lambda_2$  & 0.0140 & 0.0110 & 0.0088 & 0.0069 & 0.0053 \\
\hline
\end{tabular}
\end{center}
\caption{\emph{Navier-Stokes equations:} Effect of increasing the gap $\Delta t$ between the pair of snapshots.}\label{tab:NavierStokes2}
\end{table}
\begin{table}[!ht]
\begin{center}
\footnotesize
\begin{tabular}{|c|c|c|c|c|c|c|c|}
\hline
                             &              & $\Delta t = 0.02$ & $\Delta t = 0.01$ & $\Delta t = 0.005$ \\
\hline
\multirow{2}{*}{Clean Data}  & $\lambda_1$  & 0.9834 & 0.9688 & 0.9406 \\
\cline{2-5}
                             & $\lambda_2$  & 0.0083 & 0.0091 & 0.0104 \\
\hline
\multirow{2}{*}{$1\%$ Noise} & $\lambda_1$  & 0.8488 & 0.7384 & 0.6107 \\
\cline{2-5}
                             & $\lambda_2$  & 0.0140 & 0.0159 & 0.0217 \\
\hline
\end{tabular}
\end{center}
\caption{\emph{Navier-Stokes equations:} Effect of decreasing the gap $\Delta t$ between the pair of snapshots.}\label{tab:NavierStokes3}
\end{table}

\subsection{Fractional Equations}
\begin{figure}[!ht]
\centering
\includegraphics[width=\textwidth]{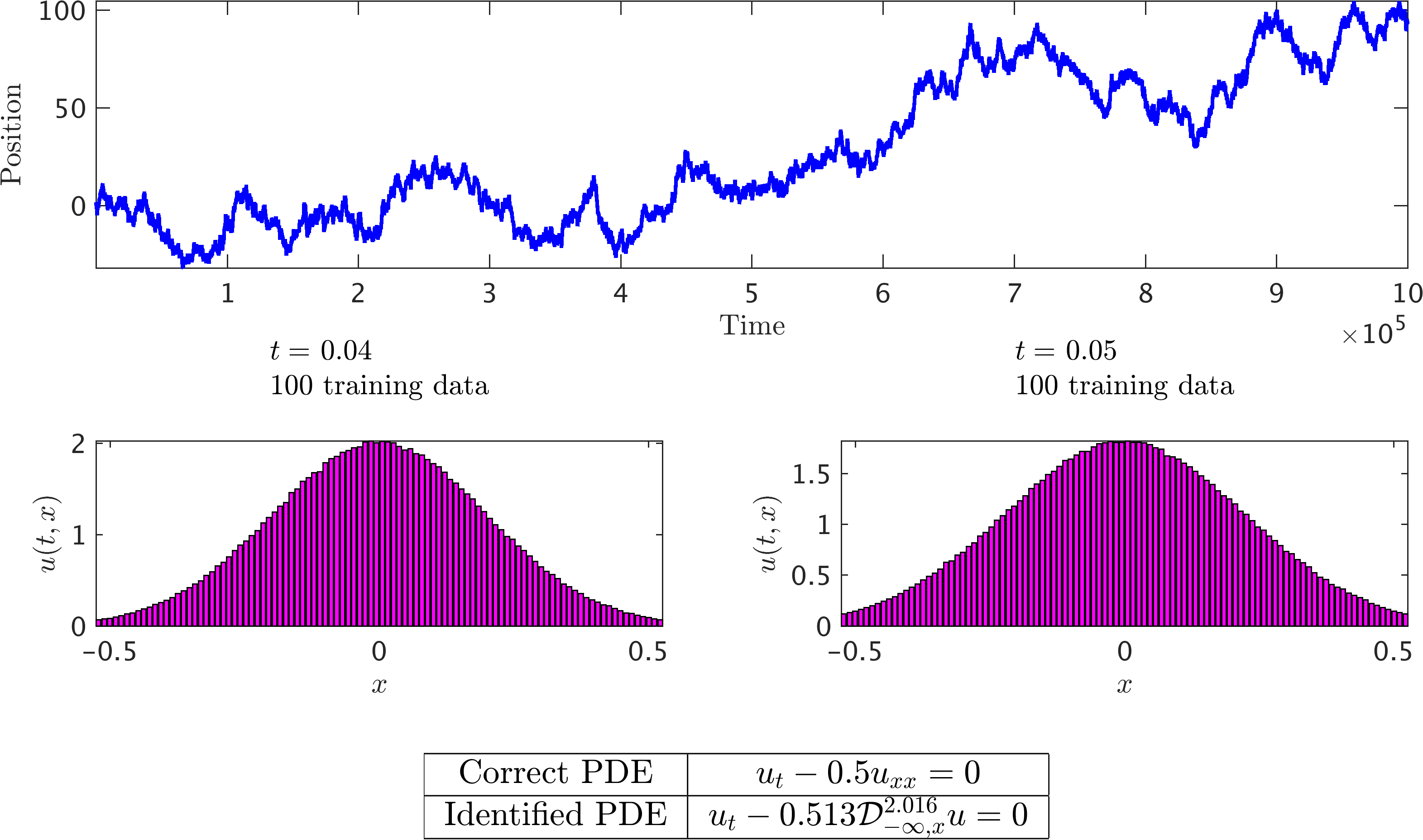}
\caption{\emph{Fractional Equation -- Brownian Motion:} A single realization of a Brownian motion is depicted in the top panel. Two histograms of the particle's displacement, being $\Delta t = 0.01$ apart, are plotted in the middle panel. The correct partial differential equation along with the identified ones are reported in the lower panel.}\label{fig:Fractional}
\end{figure}

\begin{figure}[!ht]
\centering
\includegraphics[width=\textwidth]{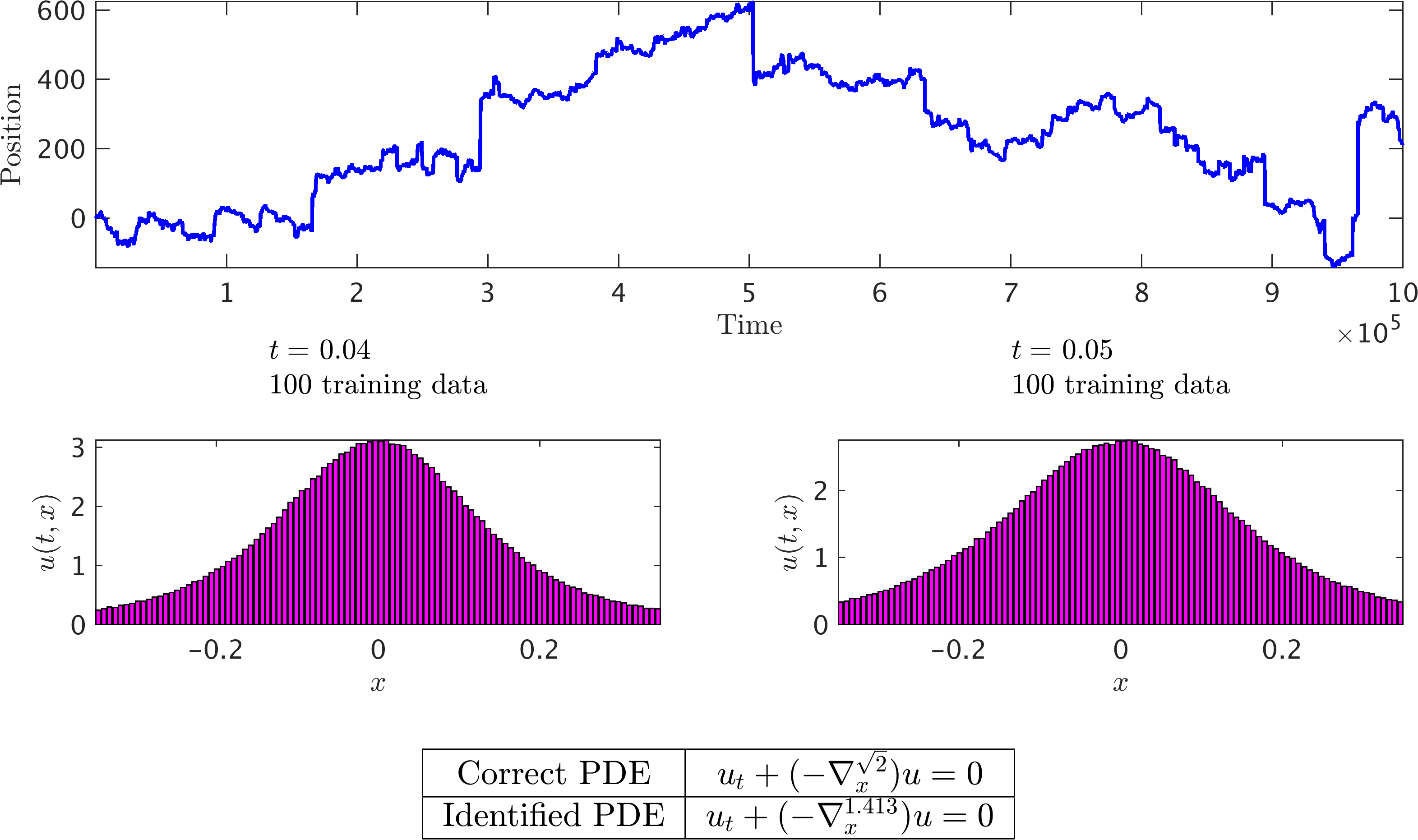}
\caption{\emph{Fractional Equation -- $\alpha$-stable L\'{e}vy process:} A single realization of an $\alpha$-stable L\'{e}vy process is depicted in the top panel. Two histograms of the particle's displacement, being $\Delta t = 0.01$ apart, are plotted in the middle panel. The correct partial differential equation along with the identified ones are reported in the lower panel.}\label{fig:Fractional_Levy}
\end{figure}

Let us consider the one dimensional fractional equation
\begin{equation}\label{eq:Fractional}
u_t - \lambda_1 \mathcal{D}^{\lambda_2}_{-\infty, x}u = 0,
\end{equation}
where $(\lambda_1, \lambda_2)$ are the unknown parameters. In particular, $\lambda_2$ is the fractional order of the operator $\mathcal{D}^{\lambda_2}_{-\infty, x}$ that is defined in the Riemann-Liouville sense \cite{podlubny1998fractional}. Fractional operators often arise in modeling anomalous diffusion processes and other non-local interactions. Integer values such as $\lambda_2=1$ and $\lambda_2=2$ can model classical advection and diffusion phenomena, respectively. However, under the fractional calculus setting, $\lambda_2$ can assume real values and thus continuously interpolate between inherently different model behaviors. The proposed framework allows $\lambda_2$ to be directly inferred from noisy data, and opens the path to a flexible formalism for model discovery and calibration. Applying the backward Euler time stepping scheme to equation (\ref{eq:Fractional}) we obtain
\begin{equation}\label{eq:FractionalBackwardEuler}
u^{n} - \Delta t \lambda_1 \mathcal{D}^{\lambda_2}_{-\infty,x} u^n = u^{n-1}.
\end{equation}
Here, $u^n(x) = u(t^n,x)$ is the hidden state of the system at time $t^n$. We make the prior assumption that
\begin{equation}\label{eq:FractionalPrior}
u^n(x) \sim \mathcal{GP}(0, k(x,x';\theta)).
\end{equation}
The prior assumption (\ref{eq:FractionalPrior}) along with the backward Euler scheme (\ref{eq:FractionalBackwardEuler}) allow us to obtain the following \emph{hidden physics model} corresponding to the fractional equation (\ref{eq:Fractional}); i.e.,
\begin{equation}\label{eq:FractionalHPM}
\begin{bmatrix}
u^{n} \\ 
u^{n-1}
\end{bmatrix}
\sim \mathcal{GP}\left(0, \begin{bmatrix}
k^{n,n} & k^{n,n-1}\\ 
k^{n-1,n} & k^{n-1,n-1}
\end{bmatrix}
\right).
\end{equation}
The only technicality induced by fractional operators has to do with deriving the kernels $k^{n,n-1}$, $k^{n-1,n}$, and $k^{n-1,n-1}$. Here, $k^{n,n-1}(x,x';\theta,\lambda_1, \lambda_2)$ was obtained by taking the inverse Fourier transform  \cite{podlubny1998fractional} of 
\[
[1 - \Delta t \lambda_1 (-iw')^{\lambda_2}]\widehat{k}(w,w';\theta), 
\]
where $\widehat{k}(w,w';\theta)$ is the Fourier transform of the kernel $k(x,x';\theta)$. Similarly, one can obtain $k^{n-1,n}$ and $k^{n-1,n-1}$. The hyper-parameters $\theta$ along with the parameters $\lambda_1$ and $\lambda_2$ are learned by minimizing the negative log marginal likelihood as outlined in section \ref{sec:Learning}. We use the hidden physics model (\ref{eq:FractionalHPM}) to identify the long celebrated relation between Brownian motion and the diffusion equation \cite{Rudye1602614}. The Fokker-Planck equation for a Brownian motion with $x(t+\Delta t) \sim \mathcal{N}(x(t), dt)$, associated with a particle's position, is $u_t = 0.5 u_{xx}$. We simulated a Brownian motion at evenly spaced time points and generated two histograms of the particle's displacement. These two histograms are $\Delta t = 0.01$ apart. As depicted in figure \ref{fig:Fractional} using only two histograms with 100 bins for each one, the algorithm is capable of identifying the correct fractional order and parameter values up to a relatively good accuracy. Moreover, let us now consider the one dimensional fractional equation
\begin{equation}\label{eq:FractionalLaplacian}
u_t + (-\nabla^\alpha_x)u = 0,
\end{equation}
where $\alpha$ is the unknown parameter and $(-\nabla^\alpha_x)$ is the fractional Laplacian operator \cite{podlubny1998fractional}. The fractional Laplacian is the operator with symbol $|w|^\alpha$. In other words, the Fourier transform of $(-\nabla^\alpha_x)u(x)$ is given by $|w|^\alpha \widehat{u}(w)$. The fractional Laplacian operator can also be defined as the generator of $\alpha$-stable\footnote{Stable distributions \cite{nolan2003stable} are a rich class of probability distributions that allow skewness and heavy tails. Stable distributions have been proposed as a model for many types of physical and economic systems.   In particular, it is argued that some observed quantities are the sum of many small terms -- the price of a stock, the noise in a communication system, etc. -- and hence a stable model should be used to describe such systems.} L\'{e}vy processes. Motivated by this observation, we simulated an $\alpha$-stable L\'{e}vy process \cite{chambers1976method, weron1995computer} and employed the hidden physics model resulting from equation (\ref{eq:FractionalLaplacian}) to identify the fractional order $\alpha$. As depicted in figure \ref{fig:Fractional_Levy} using only two histograms with 100 bins for each one, the algorithm is capable of identifying the correct fractional order up to a relatively good accuracy.

\section{Summary and Discussion} We have introduced a structured learning machine which is explicitly informed by the underlying physics that possibly generated the observed data. Exploiting this structure is critical for constructing data-efficient learning algorithms that can effectively distill information in the data-scarce scenarios appearing routinely when we study complex physical systems. We applied the proposed framework to the problem of identifying general parametric nonlinear partial differential equations from noisy data. This generality was demonstrated using various benchmark problems with different attributes. This work should be considered a direct follow up on \cite{raissi2017numerical} in which a similar methodology was employed to infer solutions to time-dependent and nonlinear partial differential equations, and effectively quantify and propagate uncertainty due to noisy initial or boundary data. The ideas introduced in these two papers provide a natural platform for learning from noisy data and computing under uncertainty. Perhaps the most pressing limitation of this work in its present form stems from the cubic scaling with respect to the total number of training data points. However, ideas such as recursive Kalman updates \cite{hartikainen2010kalman}, variational inference \cite{hensman2013gaussian}, and parametric Gaussian processes \cite{raissi2017parametric} can be used to address this limitation. 
%In terms of future work, we plan to leverage the proposed framework to study problems in control theory using state-space representations.
%
\\

Moreover, the examples studied in the current work were inspired by the pioneering work recently presented in \cite{Rudye1602614}. The authors of \cite{Rudye1602614} followed a sparse regression approach and a full set of spatio-temporal time series measurements consisting of thousands of data points. In contrast, here we used much smaller datasets with only hundreds of points and two snapshots of the systems. However, unlike the work in \cite{Rudye1602614}, here we did not use a dictionary of all possible terms involved in the partial differential equation. We could possibly include such a dictionary in our formulation but that would make our kernel evaluations more expensive. Moreover, in some systems, e.g., in an advection-diffusion-reaction system we know most of the terms of the equation, i.e., advection and diffusion but typically the reaction term is unknown. In this case, we would seek to obtain the parameters in front of the advection-diffusion and discover the functional form of the reaction term along with any parameters using the methodology outline in this paper. In comparison to \cite{Rudye1602614}, our method does not require numerical differentiation as the kernels are obtained analytically. Moreover, we do not require a regular lattice as in \cite{Rudye1602614} and can work with scattered data. An additional advantage of our approach is that it can estimate parameters appearing anywhere in the formulation of the partial differential equation while the method of \cite{Rudye1602614} is only suitable for parameters appearing as coefficients. For example, they cannot estimate the fractional order in the last example we presented in our paper or the parameters of partial differential equations (e.g., the sine-Gordon equation) involving a term like $\sin(\lambda u(x))$ with $\lambda$ being the parameter. Also, the treatment of the noise is somewhat complex in the method of \cite{Rudye1602614} as it involves some sort of filtering via e.g., singular value decomposition whereas our method can filter arbitrarily noisy data automatically via the Gaussian process prior assumptions. We believe that both methods can be used in different contexts effectively and we anticipate that this is only the beginning of a new way of thinking and formulating new and possibly simpler equations, e.g., by employing fractional operators that are naturally captured in our framework.

\section*{Acknowledgements}
This work received support by the DARPA EQUiPS grant N66001-15-2-4055, the MURI/ARO grant W911NF-15-1-0562, and the AFOSR grant FA9550-17-1-0013. All data and codes used in this manuscript are publicly available on GitHub at \url{https://github.com/maziarraissi/HPM}.



%% References


\bibliographystyle{model1-num-names}
\bibliography{sample.bib}






% \end{thebibliography}

\end{document}